# A Cognitive-based method for Intrusion Detection System

Siamak Parhizkari[1], Mohammad Bagher Menhaj[2], Atena Sajedin[2]

[1]Qazvin Islamic Azad University, 34185-1416, Qazvin, Iran.
[2]Department of Electrical Engineering, Amirkabir University of Technology, Hafez Ave., 15875-4413, Tehran, Iran.
Emails: Parhizkari.siamak@live.com, Menhaj@aut.ac.ir, Atena.sajedin@aut.ac.ir

**Abstract**

Intrusion detection is one of the important mechanisms that provide computer networks security. Due to an increase in attacks and growing dependence upon other fields such as medicine, commerce, and engineering, offering services over a network and maintaining network security have become a significant issue. The purpose of Intrusion Detection Systems (IDS) is to develop models which are able to distinguish regular communications from abnormal ones, and take the necessary actions. Among different methods in this field, Artificial Neural Networks (ANNs) have been widely used. However, ANN-based IDS encountered two main problems: Low detection precision and weak detection stability. To overcome these problems, this paper proposes a new approach based on Deep Neural Network and Support vector machine classifier, which inspired by "divide and conquer" philosophy. The proposed model predicts the attacks with better accuracy for intrusion detection rather similar methods. For our empirical study, we were taking advantage of the KDD99 dataset. Our experimental results suggest that the new approach enhance to 95.4% classification accuracy.

## 1. Introduction:

Considering the important role of Internet in our life, security of network has become the critical foundation to all web applications. Detection of Intrusion system (DIS), seek to detect the attacks of computer through scrutinizing different information records found in network processes(Anderson, 1980; Endorf et al., 2004), which can be considered as an brilliant ways to prevail with the problems in network security. The internet's intrusion compromises the data security by various internet means.

Numerous excellent studies have discussed the detection of Intrusion processes, and its role in data security. These researches have been developed in order to improve the detection stability and precision using statistical approaches, rule-based expert systems and data mining techniques. The rule-based methods compared the traffic with the rules made by the designer and examined the similarity of the patterns with the rules. But, the results of this approaches were not accurate in terms of larger datasets. Therefore, the data mining techniques were suggested. Data mining approaches attempts to gain patterns from the labeled data to empower IDS to classify data. This approach permits classifiers to recognize new attacks.

Many researchers have been used various methods in developing machine-learning approaches for designing an Intrusion Detection System, such as neural networks (Mukkamala et al., 2003), Bayesian networks, Support Vector Machines (SVM), Fuzzy Inference Systems (FISs) (Shah et al., 2004), Multivariate Adaptive Regression Splines (MARS)(Mukkamala et al., 2004). In addition, Deep learning is a strong technique that empowers computational models consists of multiple processing layers that can learn the data representations. This method is designed based on artificial neural networks (ANN) with various layers and back-propagation learning algorithm, which has been inspired by the biological neural network. However, since deep learning method is developed based on neural networks consists of more layers (deeper layers), inherently confront by local minima and malfunction in term of imbalanced datasets.

In this study, to address these problems, we suggested a modified deep learning model. To reduce the complexity and increase the accuracy, we used the clustering method. The number of clusters and the method of clustering affect the result. To find the optimum cluster numbers we focused on the trial and error route via the mean-shift clustering method. To increase the accuracy, we used SVM classifier and deep learning. The result of superimposing the methods is to increase the accuracy.

## 2. Related work

Detection of Intrusion system was proposed by Anderson(Anderson, 1980). He designed a statistical approach which analyze the user's behaviors and detect an unauthorized one. In 1987(Denning, 1987) suggested a model, based on this hypothesis that with the help of the monitor system audit record, it could detect security violations. This model had a profile to represent the behavior of users and rules to collect information about this behavior and detect intrusion. Signature-based methods have some problems like reducing applicability by confronting a small amount of data. After introducing machine learning approaches, scientists switched to this scope. Su-Yun Wu, Ester Yen in(Wu & Yen, 2009) and Animesh Patcha, Jung-Min Park in(Patcha & Park, 2007) showed some machine learning techniques like the support vector machine(SVM), decision tree, neural network, and Bayesian network. Among these techniques, the neural networks are used broadly (Ali, Al Mohammed, Ismail, & Zolkipli, 2018; Horeis, 2003; Joo, Hong, & Han, 2003; Wang et al., 2010). The number of scientists who make use of deep learning has increased in last decades. Bo Dong and Xue Wang(Dong & Wang, 2016) made a comparison between deep learning and traditional methods of intrusion detection. They found that deep learning approach proposes more accurate results, and they used oversampling to overcome to the imbalanced data.

Tuan A Tang et al.(Tang et al., 2016) offered an approach to monitor the network flow. In this approach, the authors claim that their model gained 75.75 percent accuracy. Their model consist of three hidden layers. The input and output layer have 6 and 2 dimensions, respectively. The first hidden layer has 12, the second has 6, and the third has 3 neurons. Ahmad Y Javaid et al.(Javaid et al., 2016) suggested a model based on deep learning method. They implemented flexible and effective Network Intrusion Detection (NIDS) System with Self-Taught Learning on the NSL_KDD dataset. Their model achieved 79.10 percent accuracy in 5-class classification.

Hodo et al. (Hodo, Bellekens, Hamilton, Tachtatzis, & Atkinson, 2017) introduced an extensive survey and taxonomy of deep learning and shallow learning in IDS. Many researchers reviewed the machine learning techniques, and its performance in intrusion detection, and the effectiveness of feature selection, and finally discussed the false and true positive alarm rate. Min-Joo King and Je-Won King (Kang & Kang, 2016) designed a model based on deep learning for the in-vehicle network. They claimed that in comparison with the traditional neural network, their model has revealed more accuracy. In 2017, Chauanlong Yin et al.(Yin et al., 2017) suggested a deep learning-based approach with Recurrent Neural Network (RNN). They achieved 81.29 percent accuracy on the $KDDTest^{+}$ dataset. The activation function which they utilized was Sigmoid, and the classification function was SoftMax. In ref (Shone et al., 2018) authors designed a model with deep Auto-Encoder under the name of Stacked Non-symmetric Deep Auto-Encoder. They employed the encoding portion of the Auto-encoder and reduced the dimensions and applied a random forest classifier. Their model achieved 85.42 percent accuracy.

## 3. Methodology

### 3.1 System framework

In this section, we introduced our proposed model, the dataset, and the analysis procedure. First of all, we reduced dimensions of our dataset and classified the data. Then, we trained 2 models for each cluster with Deep Neural Network (DNN) and SVM classifier and compared their result for gained the better accuracy. Next, the output of the previous step was aggregated and the results generate a simple model of neural network. Figure 1 displays our proposed model.

The idea of our model inspired from "the divide and conquer", so that after preprocessing, a dataset with Mean-Shift algorithm can be divided into the K cluster. This complexity reduction of the dataset results in reaching the higher accuracy of the model. Notably, K is not constant, and it depends on Mean-Shift implementation. After that, for each cluster, we made two models, one with DNN and the other with SVM, and then we chose the one that met more accuracy. After that, we fed each data and aggregated the result. Figure 2 illustrates the flow chart and Algorithm 1 that shows the pseudo code of our model.

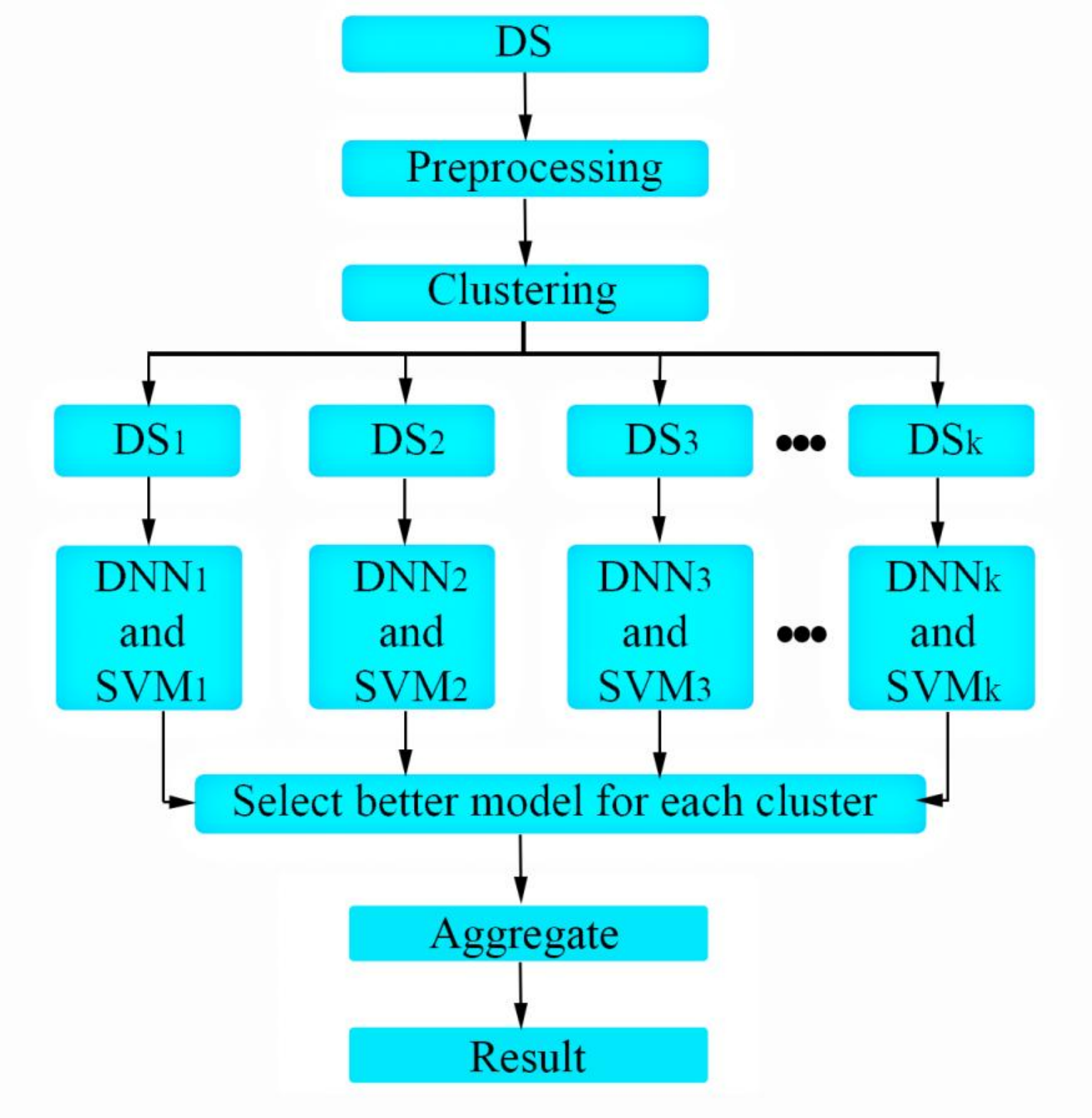

Fig 1. Model diagram

### 3.2 Dataset used for experiments

We used the NSL_KDD dataset in the experiments. The dataset contains five classes, four of them are attacks, and one of them is normal. Table 1 illustrates data. Each record of this dataset consist 41 features that describe network traffic and labeled as a specific attack. This dataset is a revised version of KDD CUP 99(Dhanabal & Shantharajah, 2015; Tavallaee, Bagheri, Lu, & Ghorbani, 2009). The classes of NSL_KDD are as follows:

- Normal: Normal traffic is something that is not classified into four attacks.
- DOS: Denial of Service is a kind of attack in which the attacker attempts to utilize the victim's resources, thereby making the victim unable to respond to a legal request.
- Probing: The attacker seeks to obtain information about network, victim and so forth.

– U2R: The attacker first logs into victim's computer via a regular account and then seek to attain the root privileges.
– R2L: The attacker attempts to invade the victim's computer without having any authority remotely.

```
function Load_dataset(dataset)
    preprocess dataset
    cluster = meanshift_cluster_dataset
    for i = 1 to len(cluster) do
        create neural_network and svm
        train neural_network and svm
        if neural_network more_accurate than svm

            selected_model = neural_network
        else
            selected_model = svm
        endif
        list[i] = selected_model(dataset)
        list[i] *= cluster[i]_grade
        augmented_data = hstack(list[i])
    endfor
    create_single_layer_neural_network
    if augmented_data is from train_set
        train single_layer_neural_network
    else
        test single_layer_neural_network
    endif
end function
```

Algorithm 1. Pseudocode of the proposed model

### 3.3 Data preprocessing

Our dataset has three nominal features; 'Protocol type' 'Service' and 'Flag, and we first labeled them to number for proper input to the neural network. We then normalize our data through Min-Max normalization method and also oversampling them in this order we reach more accurate results. Equation 1 shows the Min-Max formula in the following.

$$x_i = \frac{x_i - \text{Min}}{\text{Max} - \text{Min}} \qquad (1$$

In our dataset, we have some imbalanced data. One class like U2R has 52 records, and another class, the normal one that contains 67343 records. There are some methods to overcome this problem in the literature. One of the popular methods is oversampling. Using this method, we repeat the records of a specific class containing insufficient data to even-out them into the other classes.

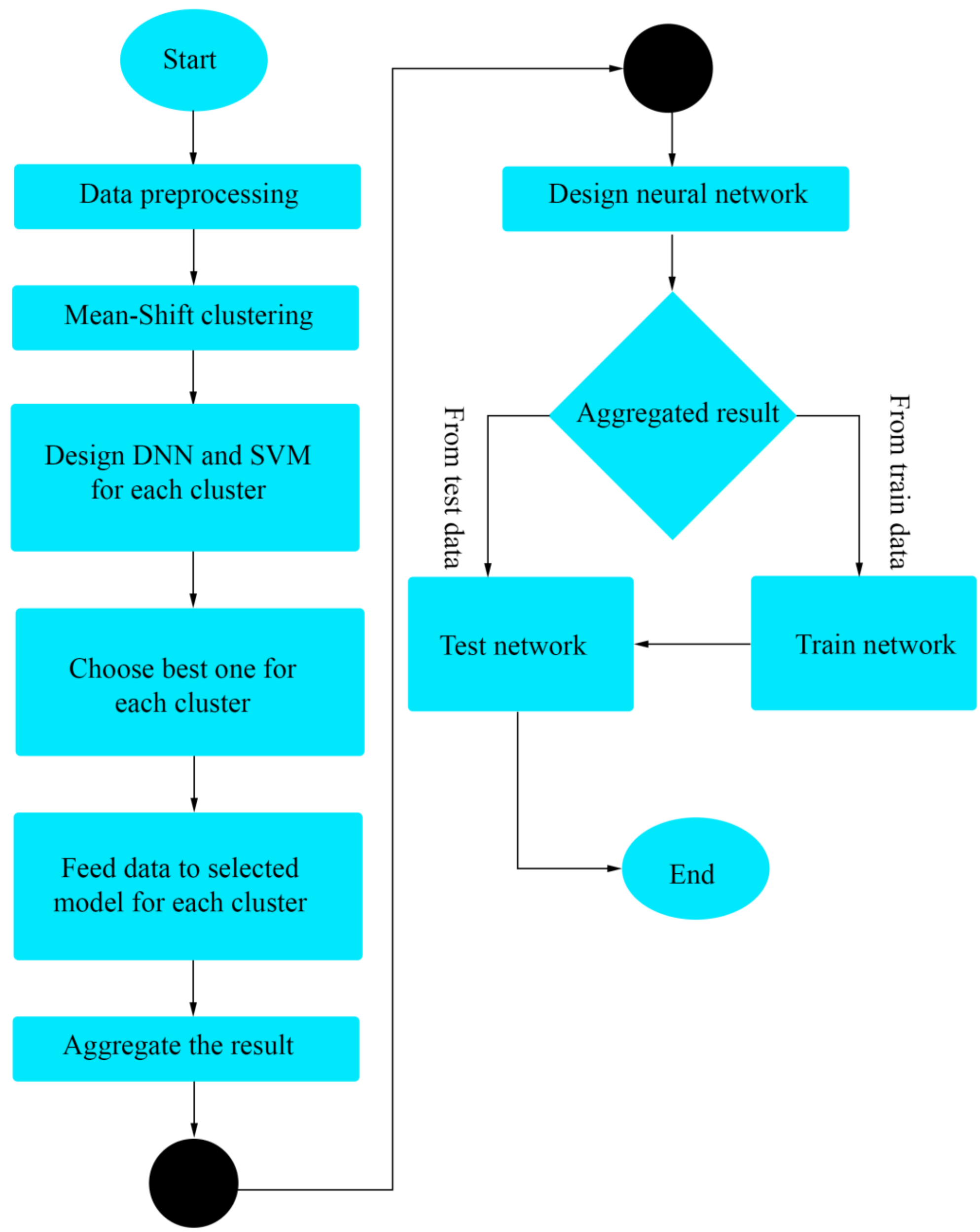

Fig 2. Flowchart of the proposed model

After preprocessing, we used auto-encoder for data dimension reduction. Each record of our dataset has 41 dimensions. With an under-complete auto-encoder, we reduced it in to 25 dimensions. Afterward, we divided our data into three distinc clusters via the mean-shift method.

### 3.5 Mean-Shift clustering

The purpose of clustering is to reduce the complexity of dataset and enhance the accuracy of our model. There are two ensembles for clustering. 1- Hard clustering. 2- Soft clustering (Bezdek, 1973). In the soft clustering, each data point can belong to more than one cluster, but in the hard one, each data belongs to one cluster. Data in the same cluster has homogeneity, and there is also heterogeneity amongst clusters. On the other hand, in terms of the methodology, there are two methods, namely hierarchical and flat ones. The hierarchical method, the number of clusters is undefinable, while as for the flat one, we can define it. The mean-shift is known as a hard and hierarchical clustering algorithm.

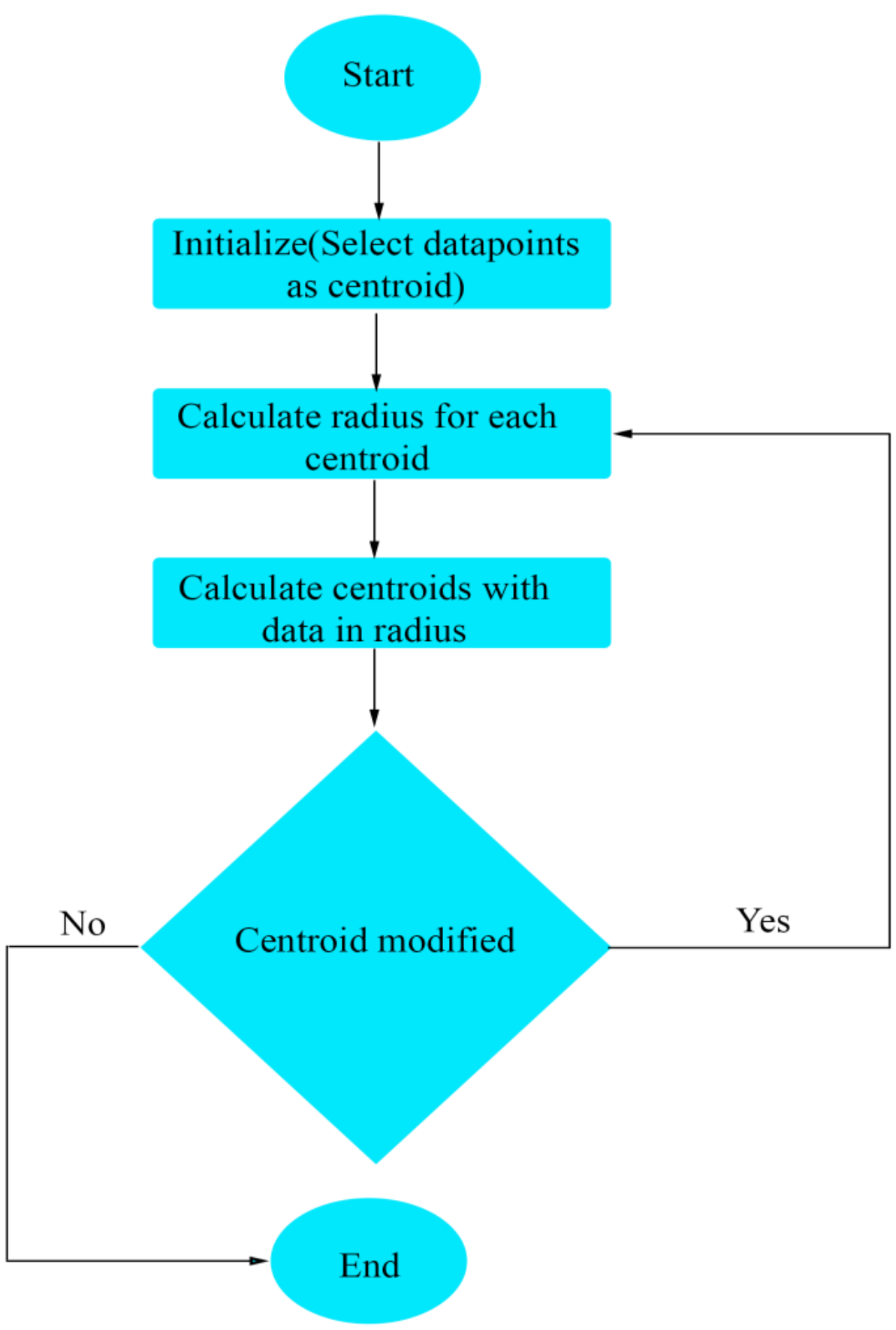

Fig 3. Flowchart of Mean-shift algorithm

First, each data point is considered centroid with a radius around it. This radius can be determined by the designer or can be calculated by a function. In the function mode, data have to shuffles, and then some of it (one-third) is selected for the calculating of its mode. The result is considered as the radius. Then the mean of each data within the radius is calculated, and the result is considered as a new centroid. These steps are iterated until there is no movement. Equation 2 shows the function that calculates the centroid.

$$m(x) = \frac{\sum_{x_i \in N(x)} k(x_1 - x)x_i}{\sum_{x_i \in N(x)} k(x_i - x)} \qquad (2$$

Where m(x) is a centroid, N(x) is the amount of data in the radius, and $k(x_1 - x)$ is Gaussian kernel. With this algorithm, our dataset is divided into 3 distinct clusters. Figure 3 shows the scheme of the mean-shift algorithm.

### 3.6 Deep Neural Network

In this section, we trained two models for each cluster and selected one, which reach better accuracy than the other. The target of these models is to learn the pattern of data in each cluster. The neural network method inspired by the brain(Menhaj, 1998). It is a network consists of many neurons in a specific and regular arrangement as many connected layers. We used the Feed-Forward neural network structured through many layers and also with back propagation learning algorithm, for updating weights of connections. Figure 4 depicts the structure of our deep neural network. The general formula of the neural network(Menhaj, 1998) is as follows:

$$a = f(wp + b) \qquad (3$$

Where (p) is the input; (w) and (b) are the network parameters (these can be modified by a learning algorithm); (f) is activation function; and (a) is output. At first (p) dot produced with (w) is aggregated with (b); then (f) applies to the result and (a) will be generated. With the algorithm of back propagation, network parameters are modified regularly, and this modification continues until the parameters are changed. Equation 3(Menhaj, 1998) shows the back propagation algorithm.

$$w_{ij}^{I}(k+1) = w_{ij}^{I}(k) - \alpha \frac{\partial \hat{F}(k)}{\partial w_{ij}^{I}(k)} \qquad (4$$

$$b_{i}^{I}(k+1) = b_{i}^{I}(k) - \alpha \frac{\partial \hat{F}(k)}{\partial b_{i}^{I}(k)}$$

$$\hat{F}(k) = \sum e_j^2(k) = \underline{e}^{T}(k) \cdot \underline{e}(k)$$

$$\underline{e}(k) = \underline{t}(k) - \underline{a}(k)$$

Where I presents I-th layer, i presents i-th element, and k means k-th epochs

Table1. the parameters

| Dataset | Normal | DOS | Probing | R2L | U2R |
|---|---|---|---|---|---|
| Training set | 67343 | 45927 | 11656 | 995 | 52 |
| Test set | 9710 | 7458 | 2421 | 2754 | 200 |

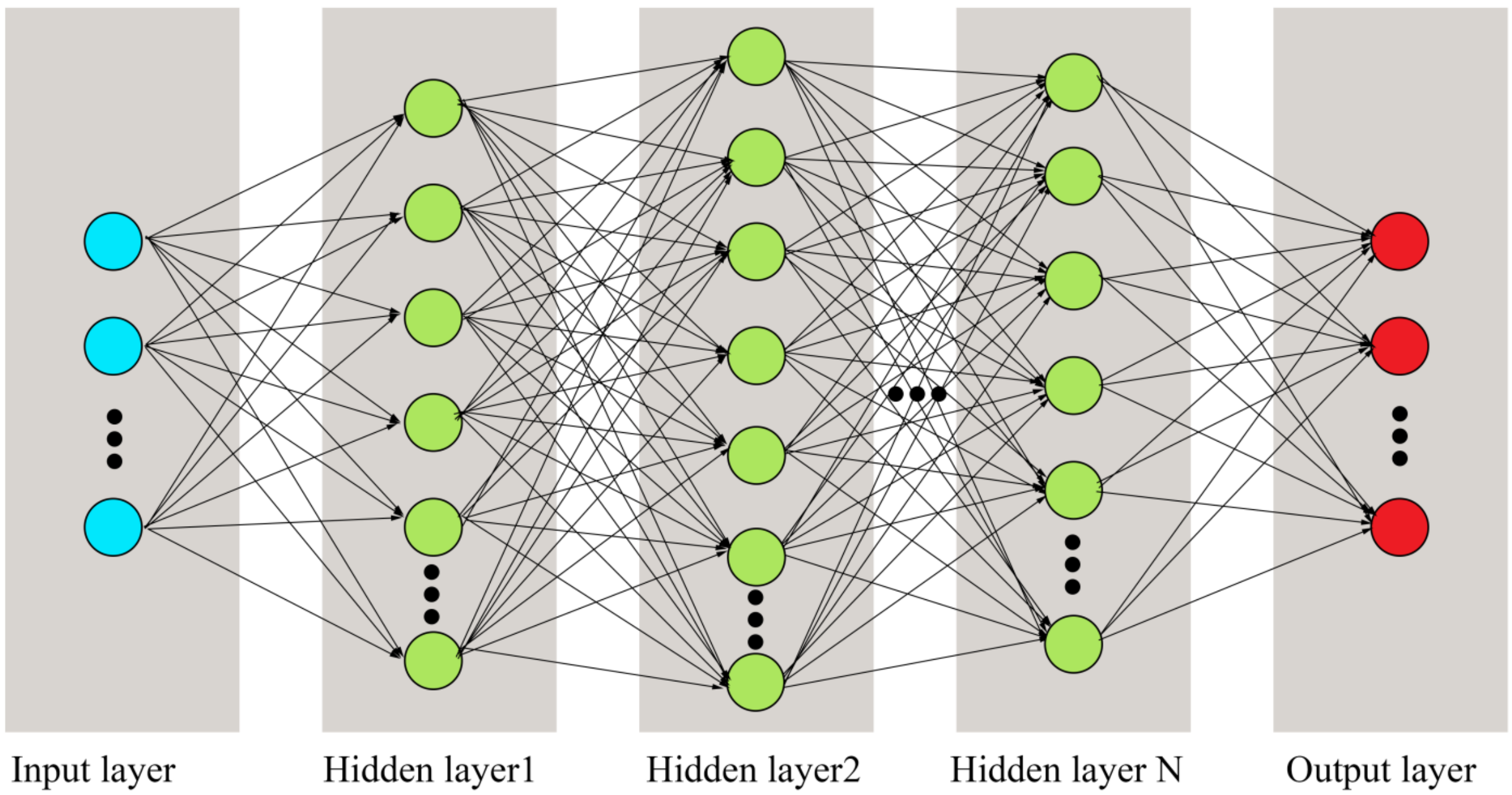

Fig 4. The network structure

### 3.7 SVM classifier

SVM(Ertekin, 2009; Vapnik, 1995) is developed to addressed the regression and classification problems. In this method, finding an optimal boundary is essential. The question arises is whether each boundary is good, what the best boundary is, and whether is it sufficient to simply separate the data. These are some critical questions which had led to the invention of SVM. There are many approaches to separating two classes of data points. The purpose of these approaches is to find a hyper plane that has the maximum margin with data on the two classes. Maximization provides this opportunity to classify data with high confidence. The hyperplane is a decision boundary that helps classify the data. The support vectors are the datapoints that affect the position and orientation of hyperplane and are close to it. With this datapoints, we can maximize the margin.

$$\theta^T x + \theta_0 \qquad (5$$

Considering equation 5, we define hyperplanes (H) as following:

$$\begin{aligned} \theta^T x^t + \theta_0 &\geq +1 \text{ when } y_i = +1 \\ \theta^T x^t + \theta_0 &\leq -1 \text{ when } y_i = -1 \end{aligned} \qquad (6$$

$H^1$ and $H^2$ are the planes:

$$H^1: \theta^T x^t + \theta_0 = +1 \quad (7$$
$$H^2: \theta^T x^t + \theta_0 = -1$$

According to the equations6, the data must be classified in the right class and must have a distance equal to or more than +1 for the positive class and equal to or less than -1 for the negative class. This is like considering margin for hyperplane.
We can rewrite equations 7 as follows.

$$y^t(\theta^T x^t + \theta_0) \geq +1 \quad (8$$

With Equation 9, the distance between each data with hyperplane can be calculated as follow:

$$\frac{|\theta^T x + \theta_0|}{||\theta||} \geq \rho \rightarrow |\theta^T x + \theta_0| \geq \rho||\theta|| \quad (9$$
$$\rho = \frac{|1 - b|}{||\theta||} - \frac{|-1 - b|}{||\theta||} = \frac{2}{||\theta||}$$

Where ρ presents a desire distance between data and hyperplane. To maximize this, we must minimize denominator$||\theta||$.

$$\min \frac{1}{2} ||\theta||^2 \quad (10$$

To solve Equation 10, we must clear the constraint that exists. To do this, we used Lagrangian formula:

$$\min_{\theta,x} L(\theta, \theta_0, \alpha) \equiv \frac{1}{2} ||\theta||^2 - \sum_{t=1}^{1} \alpha^t y^t (\theta x^t + \theta_0) + \sum_{t=1}^{1} \alpha^t \quad (11$$

**3.8 Aggregation**

Aggregation module includes three steps explained below (see Figure 5)

1. We used the dataset as input to each models.
2. The output of each model multiplies to the member degree of its cluster that is calculated from the mean-shift step (because we use the hard cluster, the member degree for each data is 0 or 1).
$$Y_{input} = [y_1. U_1, y_2. U_2, \dots, y_n. U_n]$$
Where (y) is the output from the last step, (U) is the member degree for each cluster
3. If (Y) is generated from training data, we will use it to train a simple neural network, and if it is generated from test data, we will use it to test our model.

## 4. Experimental Results

Here, we used keras("keras,") with the tensor flow backend to evaluate the performance of our proposed model. for all analysis we used HP-Envy-DV6 system, configuration: Intel corei7 3630QM CPU @ 2.4 GHz, 8 GB memory.

To perform our evaluation, we used the NSL-KDD dataset.

*Evaluation Criteria*

The following measurements were used to evaluate detection precision of IDS.

True Positive (TP): indicates whether an attack is happened and the model is detected it and classified it correctly.

True Negative (TN): shows whether the data was normal and the model detected it and classified it correctly.

False Positive (FP): indicates whether the data was normal, the model recognized it as an attack and the data was classified incorrectly.

False Negative (FN): indicates whether attack was recognized as normal data incorrectly.

With these measurements, we can calculate the other measurements:

$$\text{Accuracy} = \frac{\text{TP} + \text{TN}}{\text{TP} + \text{TN} + \text{FP} + \text{FN}} \quad (12$$

$$\text{Precision} = \frac{\text{TP}}{\text{TP} + \text{FP}} \quad (13$$

$$\text{Recall} = \frac{\text{TP}}{\text{TP} + \text{FN}} \quad (14$$

$$\text{f} - \text{score} = \frac{\text{Precision} \,.\, \text{Recall}}{\text{Precision} + \text{Recall}} \quad (15$$

$$\text{TPR} = \text{Recall} \quad (16$$

$$\text{FPR} = \frac{\text{FP}}{\text{TN} + \text{FP}} \quad (17$$

### 4.1 Cross Validation

In order to prevent bias and to assess how the result of statistical analysis will generalize to an independent dataset, we use cross-validation(Goodfellow, Bengio, & Courville, 2016). Moreover, we repeat our algorithm 10 times independently to prevent any accident and invalid results. Besides, the f-measure, which is an average of precision and recall, guaranties the accuracy of the result even if the distributions of the classes are not the same.

There are many methods for cross-validation. One of them is K-fold cross validation(Chollet, 2017). In this method, at first the dataset is divided into the k portion. One of these k portions is for the test and the remainder of it is for training. This method is repeated for (k) times. The average of the (k) validation score is the model validation score. Fig 6 displays the k-fold. In this research, we have applied the k-fold cross-validation with k=10 in the section of DNN and SVM.

### 4.2 Result

In our experiment, in the train phase, we have three clusters and six models, that selected three of them. For the first cluster, the DNN result is better than that of SVM (73% vs. 60%). This model has six hidden layers (25, 15, 15, 25, 15, and 10 neurons), as well as the output layer which contains five neurons. Sigmoid is considered to be an activation function in all layers.
The cost function is Mean Square Error, the epoch is 30, and the batch size is 64. For the second cluster, we used SVM (51% vs. 40%). In the third cluster, DNN (41% vs.22%) was chosen. This model has two hidden layers (25, 15 neurons), and the output layer has five neurons. The activation function for the hidden layers is Relu and, for the output layer, is Sigmoid, the cost function is the categorical_crossentropy, the epoch is 40, and the batch size is 128.

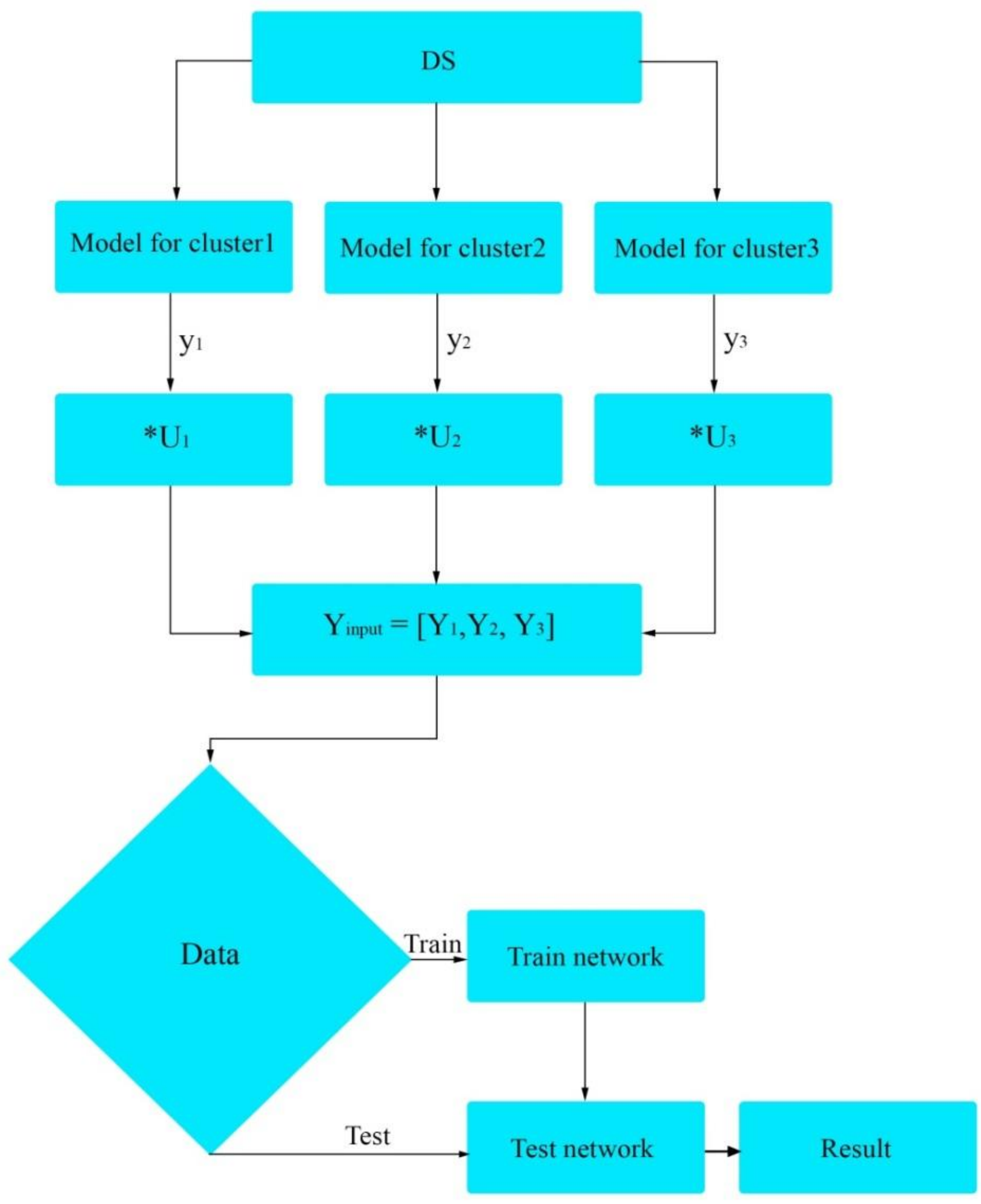

Fig 5. aggregate module

Table 2. The result of the proposed model

| Accuracy % | Precision % | Recall % | f-score % | FPR rate % | TPR rate % |
|---|---|---|---|---|---|
| **95.4** | **96.5** | **95.4** | **95.9** | **0.04** | **95.4** |

Table 3: Comparison of the accuracy of proposed model with thr related researches

| Algorithm | Accuracy % | Precision % | Recall % | f-score % | FPR rate /% | TPR rate /% |
|---|---|---|---|---|---|---|
| SMR(Javaid, Niyaz, Sun, & Alam, 2016) | 75 | 87 | 62 | 72 | - | - |
| STL(Javaid et al., 2016) | 79 | 84 | 69 | 76 | - | - |
| SDN(Tang, Mhamdi, McLernon, Zaidi, & Ghogho, 2016) | 75.75 | - | - | - | - | - |
| MLP(Moradi & Zulkernine, 2004) | 90.3 | - | - | - | - | - |
| Naïve Bayes(Miller, 2018) | 75.3 | - | - | - | - | - |
| Neural Network(Miller, 2018) | 77.8 | - | - | - | - | - |
| SVM(Miller, 2018) | 76.9 | - | - | - | - | - |
| K-means(Miller, 2018) | 74 | - | - | - | - | - |
| RNN(Yin, Zhu, Fei, & He, 2017) | 81.29 | - | - | - | - | - |
| DBN(Shone, Ngoc, Phai, & Shi, 2018) | 80.58 | 88.10 | 80.58 | 84.08 | 19.42 | - |
| S-NDAE(Shone et al., 2018) | 85.42 | 100 | 85.42 | 87.37 | 14.58 | - |
| **Proposed model** | **95.4** | **96.5** | **95.4** | **95.9** | **0.04** | **95.4** |

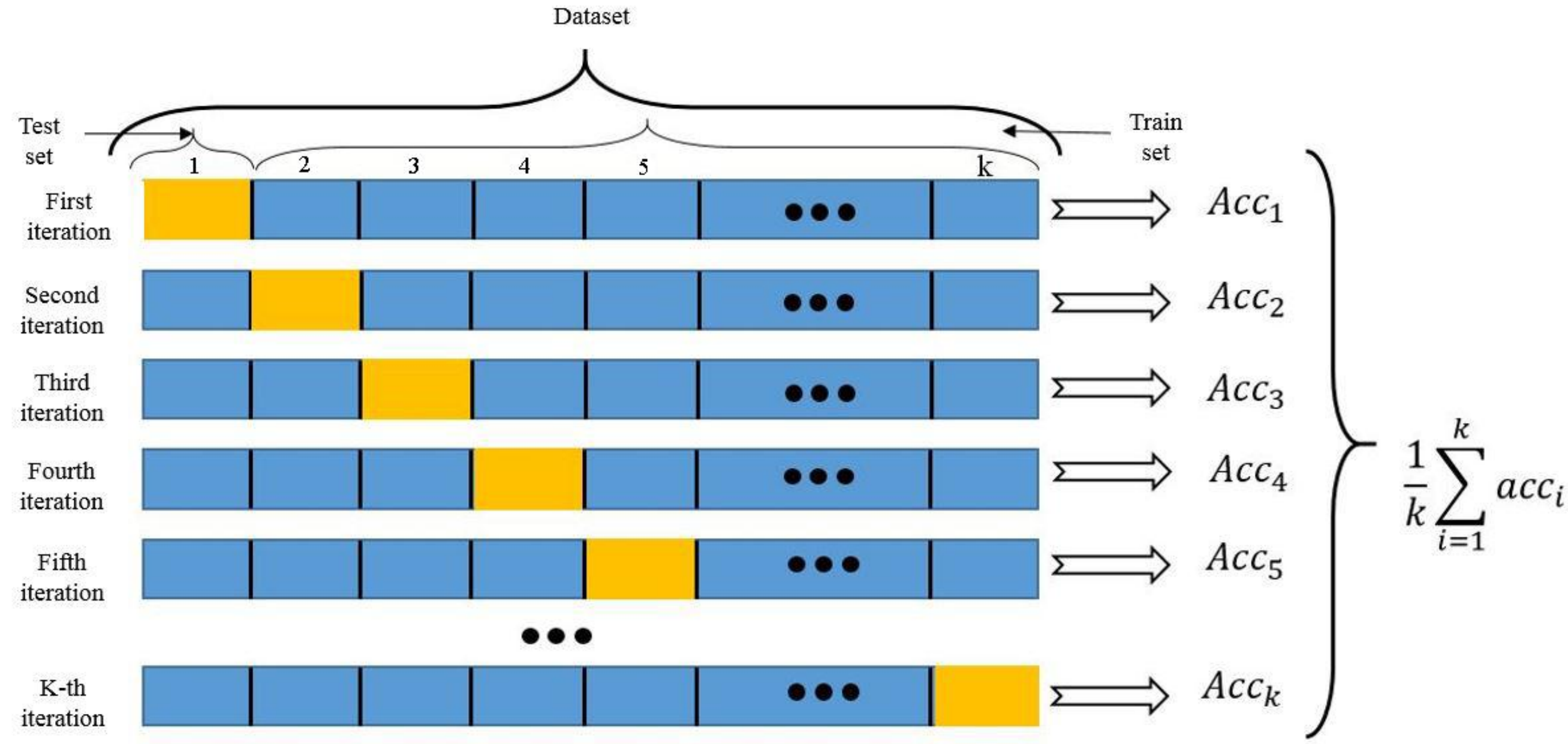

Fig.6. k-fold cross validation

In the test phase, we have three clusters. The selected model for the first cluster is DNN (77% against 67%) with six hidden layers (25, 15, 15, 25, 15, and 10 neurons) and the output layer of five neurons. To test these models, we have used all test data for each one. Similarly, the activation function for all layers is Sigmoid, the cost function is Mean Square Error, the epoch is 40, and the batch size is 64. The selected model for the second cluster is DNN (29% vs. 22%). This model has six hidden layers (25, 15, 15, 25, 15, and 10 neurons) and the output layer with five neurons. The activation functions for the hidden layers and the output layer are Relu and Sigmoid, respectively. The cost function is categorical_crossentropy, the epoch is 40, and, the batch size is 128. The third cluster model is SVM (38% vs. 29%).

For both phases, we have separately aggregated the output of each selected models. This new data is used for the next step in the aggregation section. Then, we use a simple single layer (5 neurons) of a neural network to classify the data into five categories, four attack classes, and one normal class. The activation function is Tanh, the cost function is the Mean Square error, the epoch is 30, and the batch size is 512.

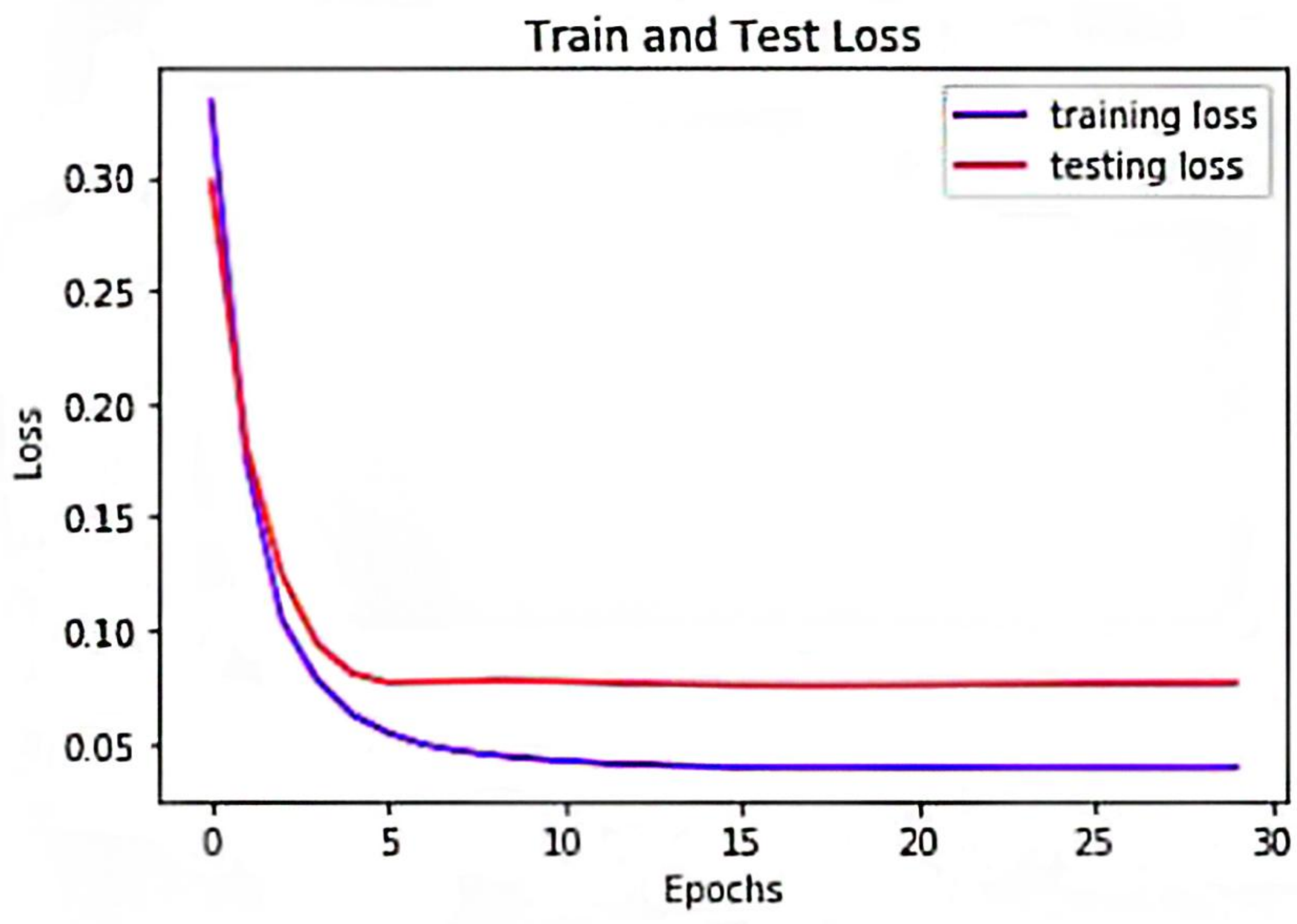

Fig 7. Train and test loss

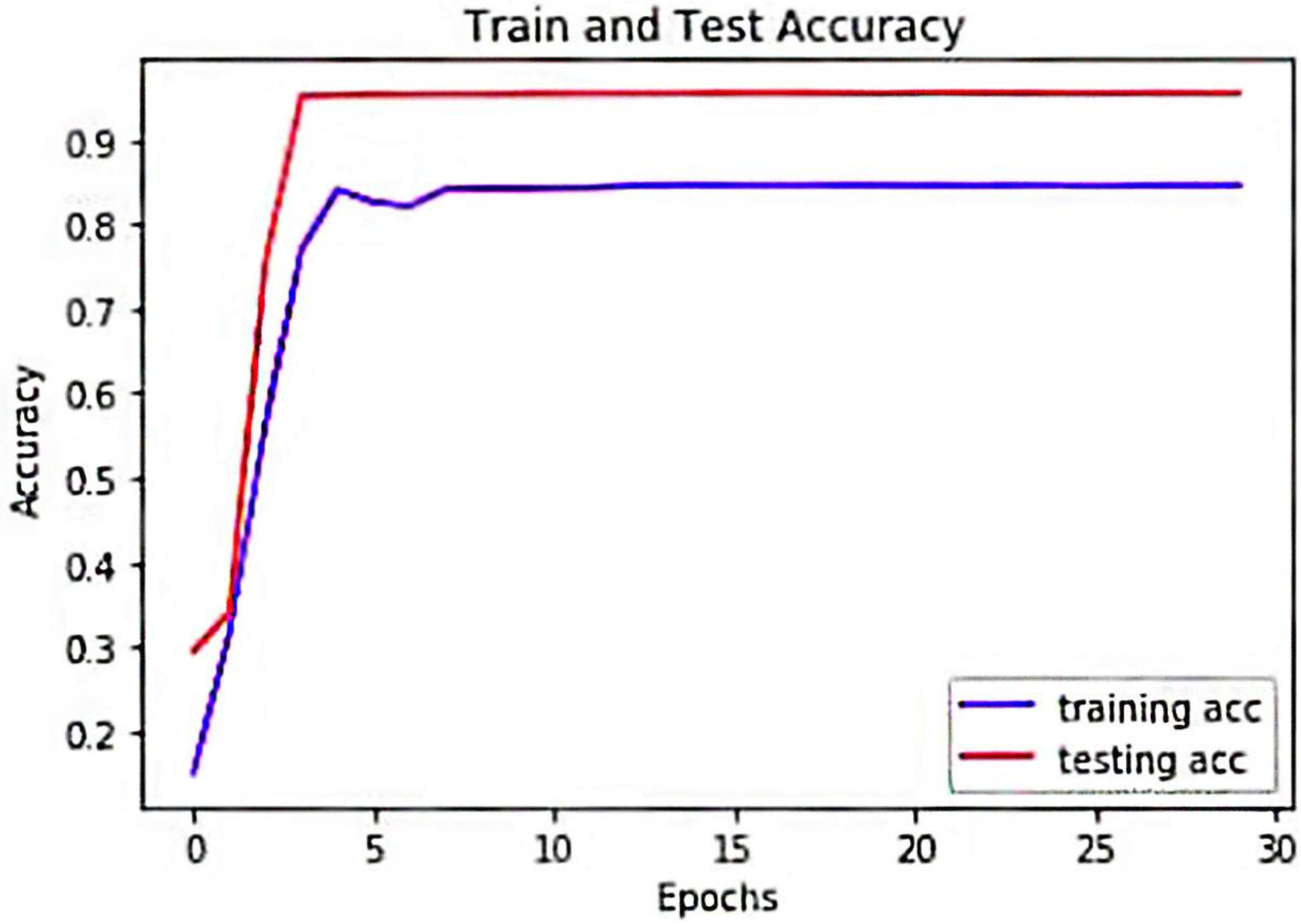

Fig 8. Train and test accuracy

We tabulated the score for different evaluation criteria for the proposed model(see table 2). In Table 3, we compared these results with some important methods in this area. In addition, Figures 7 and 8 are show the training and testing loss and accuracy, respectively.

## 5. Conclusion

One of the initial mechanisms which provide security for the computer networks is intrusion detection. Nowadays, the computer networks have to equipped with intrusion detection. The tendency of giving services over networks has been increased in various fields such as medicine, commercial, and engineering and by knowing the fact that the number of attacks has been raised, securing networks has become an important issue.

Therefore, creating the models with capability to differentiate between regular communications and atypical ones, and took the necessary steps are the major purposes of the Intrusion Detection Systems (IDS). Although Artificial Neural Networks (ANNs) have been utilized more than the other methods in this field, IDS based on ANN encountered with 2 main disadvantages which are low detection precision and weak detection stability. In this paper, a new approach based on Deep Neural Network (DNN) suggested to conquer these problems. The general mechanism of our model is as follows:

At first some of the data in the dataset was labeled properly, afterward the dataset was normalized with a Min-Max normalizer in order to fit in the limited domain. Then the data dimension was reduced in order to decrease the amount of both redundant data and computational costs. After the preprocessing part, the Mean-Shift clustering algorithm was used to create different subsets and reduce the complexity of the dataset. By taking each subset into account, the two models were trained by the Support Vector Machine (SVM) and deep learning methods.

In this research, from the two given models for each subset, the one which showed higher accuracy was chosen. The philosophy of "divide and conquer" played a significant role in promoting this idea. Therefore, it can be said that DNN is able to learn each subset rapidly. Ultimately, one way that could be used to diminish the previous step error is to train an ANN model to obtain and utilize the results so that it can anticipate the attacks. Through doing this approach, we reached 95.4 percent of accuracy.

The presented model was evaluated through different metrics such as the precision, accuracy, TPR rate, FPR rate, f-measure, and recall. Moreover, it performance were compared with methods like STL, SDN, MLP, Naïve Bayes, Neural Network, SVM, RNN, K-means, DBN, S-NDAE, and SMR. Notably, some of these mentioned methods did not contain all the information of comparative metrics in this study. According to the information which was given about the methods, it was concluded that the presented method in this study performs better than the other ones, except S-NDAE, which worked better regarding precision. But, the other methods showed significantly lower results in term of accuracy in comparison with the proposed method in this research. Interestingly, the S-NDAE method displayed 85.42 percent as to accuracy, while our method showed 95.4 percent.

## Competing interests

Te authors declare no competing interests.

## The References